\documentclass[sigconf]{acmart}
\AtBeginDocument{}
\setcopyright{acmlicensed}
\copyrightyear{2018}
\acmYear{2018}
\usepackage{booktabs}  
\usepackage{multirow}
\usepackage{enumitem}
\usepackage{graphicx}
\usepackage{tabularx}
\usepackage{booktabs}
\usepackage{booktabs}
\usepackage{multirow}
\usepackage{graphicx}
\usepackage{amsmath}
\usepackage{colortbl}
\usepackage{float}
\usepackage{xcolor}

\usepackage{booktabs}
\usepackage{multirow}
\usepackage{graphicx}
\newcolumntype{Y}{>{\centering\arraybackslash}X}

\acmDOI{XXXXXXX.XXXXXXX}
\acmConference[Conference acronym 'XX]{Make sure to enter the correct
  conference title from your rights confirmation email}{June 03--05,
  2018}{Woodstock, NY}
\acmISBN{978-1-4503-XXXX-X/2018/06}

\begin{document}
\title{GeoDecider: An Evidence-Grounded Agent for Geological Interpretation via Deliberative Reasoning}

\author{Xiaoyu Tao}
\affiliation{%
  \institution{State Key Laboratory of Cognitive Intelligence, University of Science and Technology of China, Hefei, China}
  \city{}
  \state{}
  \country{}
}
\email{txytiny@mail.ustc.edu.cn}

\author{Mingyue Cheng}
\affiliation{%
  \institution{State Key Laboratory of Cognitive Intelligence, University of Science and Technology of China, Hefei, China}
  \city{}
  \state{}
  \country{}
}
\email{mycheng@ustc.edu.cn}

\author{Jiahao Wang}
\affiliation{%
  \institution{State Key Laboratory of Cognitive Intelligence, University of Science and Technology of China, Hefei, China}
  \city{}
  \state{}
  \country{}
}
\email{jiahao.wang@mail.ustc.edu.cn}

\author{Yitong Zhou}
\affiliation{%
  \institution{State Key Laboratory of Cognitive Intelligence, University of Science and Technology of China, Hefei, China}
  \city{}
  \state{}
  \country{}
}
\email{yitong.zhou@mail.ustc.edu.cn}

\author{Qingyang Mao}
\affiliation{%
  \institution{State Key Laboratory of Cognitive Intelligence, University of Science and Technology of China, Hefei, China}
  \city{}
  \state{}
  \country{}
}
\email{maoqy0503@mail.ustc.edu.cn}

\author{Yimin Dou}
\affiliation{%
  \institution{School of Earth and Space Sciences, University of Science and Technology of China, Hefei, China}
  \city{}
  \state{}
  \country{}
}
\email{douyimin@ustc.edu.cn}

\author{Qi Liu}
\affiliation{%
  \institution{State Key Laboratory of Cognitive Intelligence, University of Science and Technology of China, Hefei, China}
  \city{}
  \state{}
  \country{}
}
\email{qiliuql@ustc.edu.cn}

\author{Shijin Wang}
\affiliation{%
  \institution{State Key Laboratory of Cognitive Intelligence, iFLYTEK Research}
  \city{}
  \state{}
  \country{}
}
\email{sjwang3@iflytek.com}

\author{Enhong Chen}
\affiliation{%
  \institution{State Key Laboratory of Cognitive Intelligence, University of Science and Technology of China, Hefei, China}
  \city{}
  \state{}
  \country{}
}
\email{cheneh@ustc.edu.cn}

\renewcommand{\shortauthors}{Tao et al.}

\begin{abstract}

Geological interpretation infers subsurface properties and structures from indirect geophysical observations. Well-log classification provides a measurable setting by assigning geological classes to depth-indexed petrophysical records. The task is difficult because different subsurface units may exhibit similar logging responses, whereas accurate interpretation often depends on local measurements, depth-wise context, domain knowledge, and reasonable transitions between neighboring layers. Existing automated methods mainly follow fixed prediction pipelines, leaving little room to gather additional evidence for difficult samples. In this work, we propose GeoDecider, an evidence-grounded agent for deliberative geological interpretation. GeoDecider retains efficient numerical prediction as the first stage, then selectively invokes tool-assisted reasoning for difficult intervals. A lightweight classifier produces point-wise predictions and estimates sample difficulty from its prediction scores. High-difficulty points act as routing anchors, triggering interval-level analysis so that nearby observations can be examined together. For each activated interval, specialized tools build an Evidence Profile that summarizes geological knowledge, depth-wise trends, previous predictions from the same well, and similar cases retrieved from training wells. Three complementary scientific views generate candidate interpretations. GeoDecider compares their supporting evidence, resolves disagreements, then applies geology-informed checks on continuity, boundary cues, and petrophysical consistency. Experiments on four public well-log benchmarks show that GeoDecider consistently outperforms representative baselines, demonstrating the value of selective evidence gathering and deliberative reasoning for geological interpretation.~\footnote{Our code is available at \url{https://github.com/Xiaoyu-Tao/GeoDecider}}

\end{abstract}

\keywords{Geological interpretation, lithology classification, scientific reasoning, large language models, tool-augmented reasoning}

\maketitle




\begin{figure*}[t]
	\centering
	\includegraphics[width=1\linewidth]{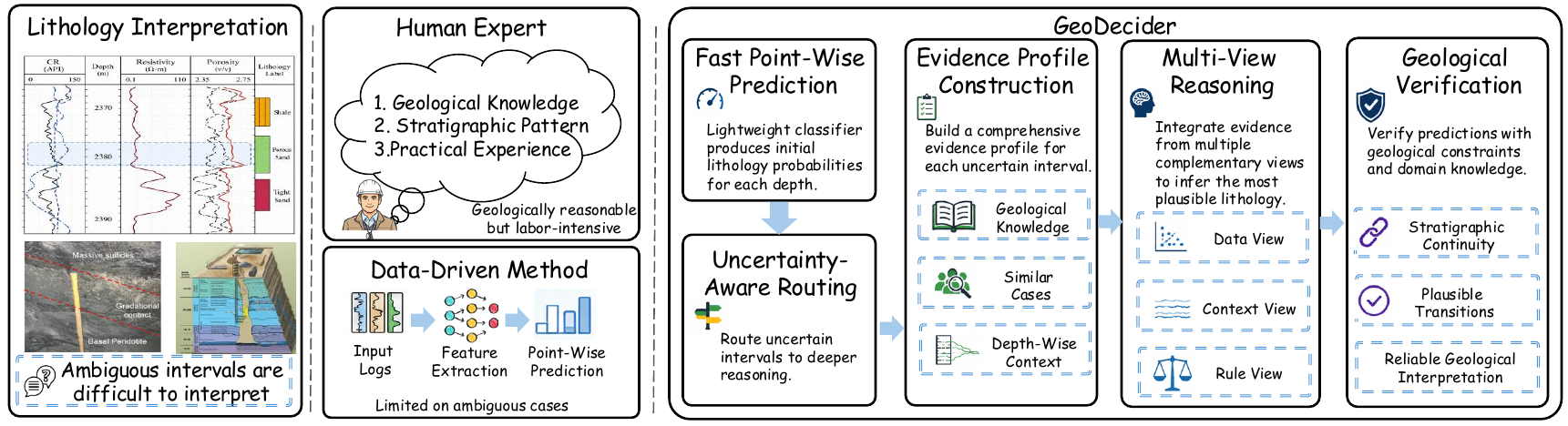}
	\caption{Overview of the geological interpretation task and the motivation for GeoDecider.}
	\label{data_pic}
	\Description{A depth-indexed multivariate well-log example with logging curves and corresponding lithology labels.}
\end{figure*}

\section{Introduction}
Geological interpretation infers subsurface properties and stratigraphic organization from indirect observations, supporting reservoir evaluation and geological modeling~\cite{schmidt2009distilling,chen2025geological,beloborodov2024automated}. Well-log target-class interpretation provides a concrete setting by assigning rock types, lithofacies, or geological layers to depth-indexed measurements~\cite{jumper2021highly,liu2025logging}. Despite its supervised classification formulation, the task cannot be reduced to independent point-wise prediction. Similar logging responses may correspond to different subsurface units, making accurate interpretation dependent on local measurements, depth-wise context, domain knowledge, and stratigraphic constraints~\cite{saleem2025multiple}. We therefore formulate automated well-log classification as an evidence-grounded interpretation problem that requires both point-level accuracy and depth-wise coherence.

Over the years, well-log interpretation has evolved from expert-driven analysis to data-driven methods~\cite{shi2023refined,wu1987automated}. As shown in Figure~\ref{data_pic}, traditional expert interpretation integrates well-log observations with geological knowledge and stratigraphic patterns but remains labor-intensive and hard to scale. Machine learning and deep learning models have improved automated lithology and facies classification~\cite{dong2023deep,hall2016facies}. Recent studies further explore advanced deep models. Sequence models and stratigraphic priors, including recurrent, CRF-based, and HMM-based approaches, exploit depth-wise correlations to improve sequence continuity~\cite{park2022data,liu2021lithological,schumann2002hidden}. However, these methods typically follow fixed inference pipelines: they cannot determine when further analysis is needed, acquire complementary evidence, or reconcile conflicting interpretations through verification~\cite{li2025large}. Thus, ambiguous intervals, formation boundaries, and overlapping logging responses remain challenging, motivating conditional and evidence-grounded workflows rather than more complex single-pass predictors.
Recent large language models (LLMs) offer opportunities for integrating domain knowledge, tool usage, and hypothesis comparison in scientific interpretation~\cite{chang2023llm4ts,tan2024language}. Tool-augmented and agentic reasoning further suggest that LLMs can coordinate external evidence and intermediate decisions rather than acting as static predictors~\cite{wang2023describe}. However, applying LLMs directly to raw numerical well logs is inefficient and insufficiently grounded. The key challenge is coordinating efficient prediction with selective evidence acquisition and deliberative reasoning while maintaining geological consistency.

To address this challenge, we propose GeoDecider, a scientific agent for geological interpretation via evidence-grounded deliberation. As shown in Figure~\ref{data_pic}, we regard GeoDecider as a scientific agent because it detects observations that warrant further investigation, conditionally activates specialized tools, compares competing interpretations through separate scientific views, and records the evidence used to verify the resulting decision. GeoDecider transforms conventional single-pass well-log target-class classification into an adaptive fast-slow interpretation workflow. A lightweight classifier produces point-wise predictions and estimates sample difficulty from its prediction scores. Rather than treating each high-difficulty point as an isolated query, GeoDecider uses such points as routing anchors that activate deliberation over their containing depth intervals. For each activated interval, it collects geological knowledge, depth-wise trends, analogical cases retrieved from training wells, and finalized predictions from preceding depths, organizing them into a provenance-preserving Evidence Profile. Data-, context-, and knowledge-oriented views then produce candidate sequences, and an evidence-aware reconciliation stage resolves their agreements and conflicts. Finally, GeoDecider applies geology-informed checks based on stratigraphic continuity, coordinated log changes, and petrophysical compatibility to verify and, when necessary, refine the interval-level interpretation.

Our contributions are summarized as follows:
\begin{itemize}[leftmargin=*]
\item We propose GeoDecider, a scientific agent transforming well-log classification into an adaptive fast-slow workflow. Prediction scores identify high-difficulty routing anchors, activating evidence-grounded deliberation over depth intervals.

\item  
We develop a deliberative reasoning workflow that builds an Evidence Profile through specialized tools, reconciles candidate sequences from complementary views, and verifies revisions using continuity, boundary, and petrophysical evidence.
\item  
Experiments on public well-log benchmarks show that GeoDecider outperforms baselines. Analyses reveal depth-wise geological context as key evidence for ambiguous intervals, highlighting the need for contextual evolution beyond isolated measurements.

\end{itemize}

\section{Related Work}
\subsection{Lithology Interpretation from Well Logs}

Lithology interpretation from well logs aims to infer subsurface geological compositions, facies, and stratigraphic units from petrophysical measurements, supporting reservoir evaluation, geological modeling, and resource characterization~\cite{karpatne2018machine,reichstein2019deep}. Early interpretation relied on expert-driven analysis, where geoscientists integrated logging responses with domain knowledge, stratigraphic patterns, and geological experience~\cite{bressan2020evaluation,bergen2019machine}. Although physically meaningful, such approaches require substantial manual effort and are difficult to scale to large volumes of well-log data~\cite{jiang2024machine,nguyen2025classifying}.
With the development of machine learning, data-driven approaches have been widely adopted for automated lithology classification. Traditional methods learn patterns from engineered logging features, while deep learning models extract nonlinear representations from raw or multi-dimensional measurements~\cite{ruiyi2021lithology,rathore2023well}. These approaches improve prediction efficiency but typically formulate interpretation as a direct mapping from observations to labels, making them vulnerable to  ambiguous responses~\cite{fan2025logging,pang2025empowering,liang2024foundation}.
To capture the sequential nature of geological formations, subsequent studies incorporated depth-wise dependencies and geological constraints through sequence models and structure-aware approaches, including recurrent networks, probabilistic graphical models, and stratigraphic consistency modeling~\cite{park2022data,liu2021lithological,schumann2002hidden}. While improving continuity and geological plausibility, these methods rely on predefined structures or manual constraints, limiting their ability to adaptively acquire evidence, compare interpretations, and resolve uncertainty through reasoning~\cite{kusuma2025leveraging}. Therefore, effective lithology interpretation requires not only accurate classification but also evidence-grounded reasoning for dynamic decision refinement.

\subsection{Evidence-Grounded LLM Reasoning}
Large language models (LLMs) have demonstrated strong capabilities in language understanding, knowledge integration, and reasoning~\cite{zhang2024large,jiang2024empowering,jin2024position}. Unlike conventional predictive models, LLM-based reasoning decomposes problems into intermediate inference processes and generates decisions conditioned on contextual information. Chain-of-thought (CoT), self-consistency, and related structured prompting strategies enable multi-step reasoning, while retrieval-augmented generation improves knowledge grounding~\cite{wei2022chain,wang2022self,nakano2021webgpt}.
Recent agentic reasoning frameworks further extend LLMs by enabling tool interaction and external information retrieval~\cite{yao2023react,schick2023toolformer}. By integrating specialized models and computational tools, LLMs can perform dynamic problem solving beyond static generation.
However, applying LLM reasoning to scientific interpretation remains challenging due to heterogeneous observations, domain constraints, and the need for traceable evidence~\cite{xu2024geopredict,luo2025time}. Existing approaches mainly target textual tasks and lack mechanisms for integrating structured measurements, domain tools, and physical consistency~\cite{zhang2025unleashing,mohammadi2025evaluation}. Reliable geological reasoning requires decisions grounded in observable evidence and geological principles.
Motivated by these challenges, we develop evidence-grounded reasoning that combines prediction, evidence acquisition, and geological verification. LLMs serve as reasoning agents to integrate evidence and refine ambiguous interpretations.

\section{Preliminaries}
This section defines depth-indexed well-log sequences, summarizes the geological characteristics that motivate contextual reasoning, and formalizes the interpretation task studied in this work.

\subsection{Well-Log Sequences and Labels}

A well log consists of multivariate petrophysical measurements sampled at ordered depth locations $\mathbf{d}=[d_1,\ldots,d_L]$, where $d_1<\cdots<d_L$. For a well with $L$ depth points and $M$ logging channels, let $\mathbf{x}_t\in\mathbb{R}^{M}$ denote the measurements at depth $d_t$. The complete well-log sequence is represented as $\mathbf{X}=[\mathbf{x}_1^\top;\ldots;\mathbf{x}_L^\top]\in\mathbb{R}^{L\times M}$, where the $t$-th row of $\mathbf{X}$ corresponds to $\mathbf{x}_t^\top$. Typical channels include gamma ray, resistivity, density, neutron porosity, and sonic measurements, while available channels vary across datasets. The corresponding target interpretation sequence is $\mathbf{Y}=[y_1,y_2,\ldots,y_L]\in\mathcal{C}^{L}$, where $y_t\in\mathcal{C}$ denotes the target interpretation class at depth $d_t$, and $\mathcal{C}$ is the dataset-specific class set. Depending on the benchmark, target classes may represent rock lithologies, lithofacies, or geological layers. Lithology interpretation is the primary setting.

A \emph{sample} denotes an individual depth point, whereas an \emph{interval} denotes a consecutive group of points. For a target depth $d_t$, the contextual indices within a physical depth radius $\rho$ are defined as $\mathcal{J}_t(\rho)=\{j\mid |d_j-d_t|\leq\rho\}$, and the corresponding context sequence is $\mathbf{X}^{\mathrm{ctx}}_t=(\mathbf{x}_j)_{j\in\mathcal{J}_t(\rho)}$, where $\mathbf{X}^{\mathrm{ctx}}_t$ preserves the increasing-depth order and $\rho$ uses the same physical unit as $d_t$. A point-wise model takes $\mathbf{z}_t=\mathbf{x}_t$, whereas a context-aware model uses $\mathbf{z}_t=\phi(\mathbf{X}^{\mathrm{ctx}}_t)$ with feature extractor $\phi$.

\subsection{Geological Characteristics of Lithology Interpretation}
Lithology interpretation differs from conventional point-wise classification in three respects~\cite{wu1987automated,shi2023refined,gil2014amplify}. First, the mapping from logging responses to lithology is non-unique: different rock types may exhibit overlapping petrophysical signatures, while the same lithology may vary with porosity, fluid content, pressure, and burial conditions. Second, interpretations depend on ordered depth-wise context. Surrounding trends, turning points, and formation boundaries can provide information that is unavailable from an isolated measurement, motivating models that exploit correlations across neighboring depths. Third, unsupported isolated label changes can fragment an otherwise continuous local sequence. Local continuity is therefore a useful diagnostic, but it is only a proxy rather than proof of complete geological validity. Reliable interpretation consequently requires local pattern recognition, contextual evidence integration, and explicit checking of the predicted sequence against the observed well-log context.

\subsection{Problem Formulation}

Within the broader setting of geological interpretation, this work focuses on well-log lithology interpretation, formulated as supervised depth-wise classification of dataset-specific interpretation classes. Geological interpretation aims to infer subsurface units from indirect measurements, while target-class prediction provides the operational task for model training and evaluation.

Given labeled training wells $\mathcal{D}_{\mathrm{train}}=\{(\mathbf{d}^{(i)},\mathbf{X}^{(i)},\mathbf{Y}^{(i)})\}_{i=1}^{N}$, where $\mathbf{d}^{(i)}\in\mathbb{R}^{L_i}$, $\mathbf{X}^{(i)}\in\mathbb{R}^{L_i\times M}$, and $\mathbf{Y}^{(i)}\in\mathcal{C}^{L_i}$ denote the depth locations, well-log sequence, and interpretation sequence of the $i$-th well, the goal is to predict an unseen well's target sequence $\hat{\mathbf{Y}}=\mathcal{F}(\mathbf{X},\mathbf{d};\mathcal{D}_{\mathrm{train}},\mathcal{K})$, where $\mathcal{K}$ denotes domain knowledge and $\hat{y}_t\in\mathcal{C}$ is the prediction at depth $d_t$. The target well is disjoint from training wells, and retrieved analogical cases must originate from $\mathcal{D}_{\mathrm{train}}$.
Reliable interpretation requires both point-level accuracy and sequence-level coherence. The former measures agreement with reference classes, while the latter evaluates depth-wise consistency and avoids unsupported isolated changes. However, ambiguous samples and intervals remain challenging because local measurements may provide insufficient or conflicting evidence. Resolving such cases requires integrating surrounding trends, domain knowledge, historical predictions, and analogous observations. Therefore, we seek a framework that can selectively acquire evidence, perform traceable reasoning, and generate predictions that are both accurate and consistent with depth-wise geological context.

\begin{figure*}[t]
		\centering
		\includegraphics[width=1\linewidth]{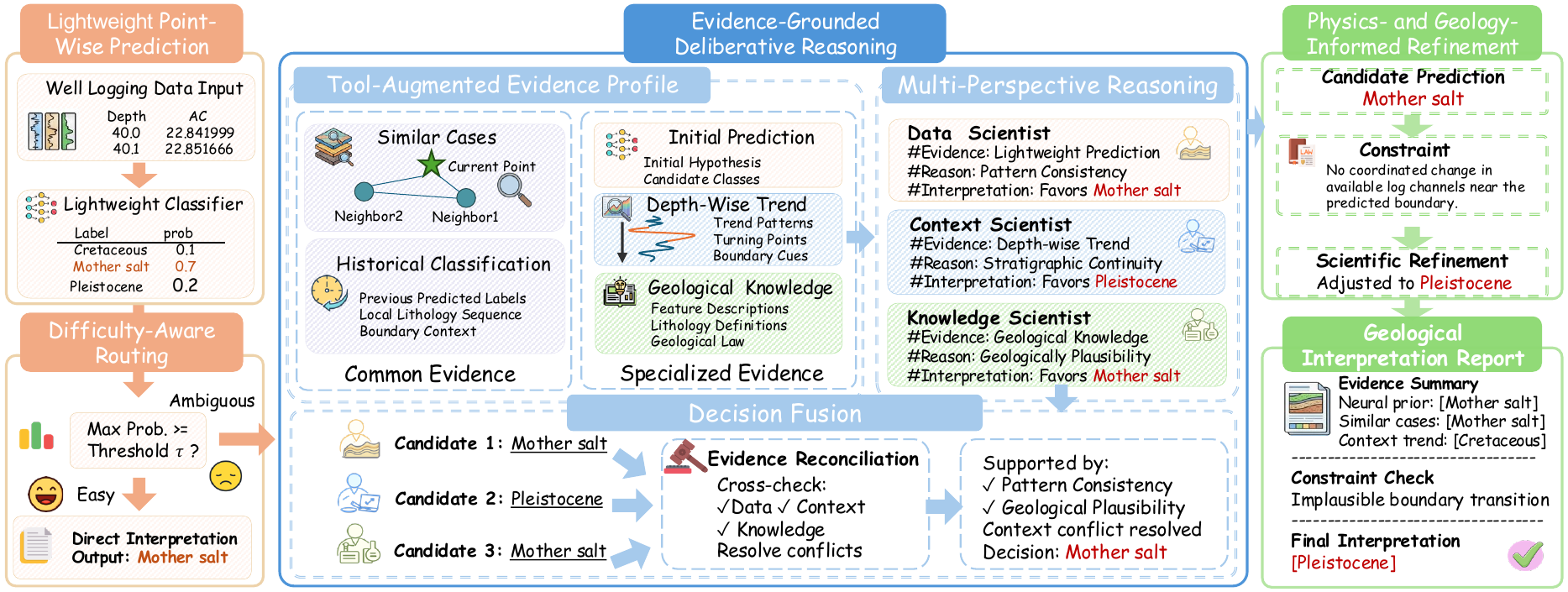}
		\caption{The GeoDecider framework. Point-wise prediction scores identify high-difficulty routing anchors that activate interval-level evidence construction, multi-perspective reasoning, evidence reconciliation, and physics- and geology-informed refinement.}
	    \Description{A fast point-wise predictor supplies initial labels and difficulty scores. Activated intervals are augmented with similar cases, previous predictions, depth-wise trends, and geological knowledge; three scientific views produce candidate sequences that are reconciled and refined into a final geological interpretation.}
		\label{framework}
\end{figure*}
\section{The Proposed GeoDecider}

GeoDecider operationalizes this formulation through a fast-slow workflow. The fast stage produces predictions and difficulty signals, while the slow stage reasons over difficult intervals. This design preserves geological context by estimating sample difficulty, constructing evidence, comparing hypotheses, and refining sequences.

\subsection{Framework Overview}
As illustrated in Figure~\ref{framework}, GeoDecider processes the target well as ordered intervals. A lightweight classifier estimates sample difficulty from prediction scores and marks high-difficulty points as routing anchors. Intervals without routing anchors retain fast predictions, while anchored intervals activate deliberation with neighboring observations. Specialized tools construct a provenance-preserving Evidence Profile, and three role-conditioned calls analyze intervals from data, contextual, and geological perspectives. Evidence-aware fusion reconciles candidate sequences, followed by refinement using continuity, log variations, and petrophysical compatibility. Intervals are processed shallow-to-deep, allowing previous predictions to provide historical evidence. No target-well labels are exposed during reasoning under strict evaluation protocols.
\subsection{Lightweight Point-Wise Prediction}
Well-log datasets contain many observations with comparatively distinctive logging patterns. GeoDecider uses a conventional classifier as the fast stage for these routine cases and as a statistical prior for subsequent reasoning. The framework is not restricted to a particular architecture: the classifier may be point-wise or context-aware, provided that it returns a score for every target class.
Given the feature representation $\mathbf{z}_t$ defined from the target measurement or its contextual window, the lightweight classifier $f_{\mathrm{fast}}$ produces a probability distribution over the candidate target classes, $\mathbf{p}_t=f_{\mathrm{fast}}(\mathbf{z}_t)$, where $\mathbf{p}_t=[p_{t,1},\ldots,p_{t,|\mathcal{C}|}]$ and $\mathcal{C}$ denotes the dataset-specific target class set. The initial prediction is obtained as $\hat{y}^{\mathrm{fast}}_t=\arg\max_{c\in\mathcal{C}}p_{t,c}$. As defined in the preliminaries, $\mathbf{z}_t=\mathbf{x}_t$ for a point-wise classifier and may be obtained from $\mathbf{X}^{\mathrm{ctx}}_t$ for a context-aware classifier. GeoDecider treats the resulting scores as model outputs rather than calibrated probabilities. Their role in the framework is to provide an initial hypothesis and a scheduling signal for deciding where additional analysis may be warranted.

\subsection{Difficulty-Aware Routing}
Well-log observations exhibit heterogeneous classification difficulty. We estimate sample difficulty from the maximum class score:
\begin{equation}
	q_t = \max_{c \in \mathcal{C}} p_{t,c}.
\end{equation}
A lower $q_t$ indicates a higher-difficulty sample because the fast classifier does not strongly prefer a single target class. Given a routing threshold $\tau$ selected on validation wells, the point-level routing indicator is:
\begin{equation}
	r_t =
	\begin{cases}
		0, & q_t \geq \tau,\\
		1, & q_t < \tau,
	\end{cases}
\end{equation}
where $r_t=1$ marks depth $t$ as a high-difficulty routing anchor. Let $\mathcal{B}_b=\{s_b,\ldots,e_b\}$ denote the indices of the $b$-th consecutive inference interval and define:
\begin{equation}
	a_b
	=
	\mathbb{I}
	\left(
	\sum_{t\in\mathcal{B}_b} r_t > 0
	\right).
\end{equation}
Here, $a_b=1$ activates deliberation for the complete interval rather than only for the routing anchors. This distinction is important because neighboring high-score observations provide context for resolving a difficult boundary and may themselves be reconsidered during interval-level reconciliation. If $a_b=0$, the fast predictions for the interval are retained without invoking the slow stage. The threshold $\tau$ is fixed using validation wells before test inference. 

\subsection{Evidence-grounded Deliberative Reasoning}
For an activated interval, GeoDecider does not directly ask an LLM to reclassify raw well-log inputs. It first constructs a structured Evidence Profile through specialized tools and then performs multi-perspective reasoning followed by evidence-aware reconciliation.
\subsubsection{Tool-Augmented Evidence Profile Construction}

For an activated interval $\mathcal{B}_b$, GeoDecider collects complementary evidence and organizes it as the ordered record:
\begin{equation}
	\mathcal{E}_b =
	\left(
		E_b^{\mathrm{pred}},
		E_b^{\mathrm{case}},
		E_b^{\mathrm{context}},
		E_b^{\mathrm{history}},
		E_b^{\mathrm{know}}
	\right),
\end{equation}
where each component contains a source identifier, content, and relevance to classes. This record preserves provenance to distinguish outputs, retrieved cases, trends, predictions, and domain knowledge.
Five components are organized into two categories: retrieved cases and historical predictions constitute common evidence, while initial predictions, depth-wise trends, and geological knowledge provide specialized evidence.

\textbf{Initial prediction evidence.}
For every $t\in\mathcal{B}_b$, the profile records the initial label $\hat{y}^{\mathrm{fast}}_t$ and the maximum prediction score $q_t$ used for difficulty estimation; the complete score vector $\mathbf{p}_t$ may also be retained when it is available from the base classifier. These quantities are treated as statistical evidence rather than direct final decisions and are further interpreted during subsequent reasoning stages to support reliable geological interpretation processes.

\textbf{Geological knowledge tool.}
The knowledge base contains feature descriptions, dataset-specific class descriptions, and geology-informed interpretation guidelines concerning attribute combinations, boundary cues, and local transitions. The knowledge source is fixed before test inference and is supplied with identifiers so that subsequent decisions can cite the rules or descriptions they use.

\textbf{Neighbor retrieval tool.}
Let $\mathcal{U}_b=\{t\in\mathcal{B}_b\mid r_t=1\}$ be the routing anchors in interval $b$. For each anchor, GeoDecider retrieves $k$ analogical observations from the labeled training wells:
\begin{equation}
	\mathcal{N}_b
	=
	\bigcup_{t\in\mathcal{U}_b}
	\operatorname{TopK}_{(i,j)\in\mathcal{D}_{\mathrm{train}}}
	\operatorname{sim}
	\left(
		\psi(\mathbf{z}_t),
		\psi(\mathbf{z}^{(i)}_j)
	\right),
\end{equation}
where $(i,j)$ identifies depth $j$ in training well $i$, $\psi$ is a feature transform fitted on training wells, and $\operatorname{sim}$ is a fixed similarity function. Each record stores well identifier, depth, target label, similarity score, and reference identifier. Thus, $E_b^{\mathrm{case}}=\mathcal{N}_b$ provides empirical analogies without accessing labels from validation or test wells~\cite{NEURIPS2024_053ee34c}.

\textbf{Depth-wise trend analysis tool.}
The trend tool analyzes the observed logging curves within $\mathcal{B}_b$ together with available measurements immediately above and below the interval. It summarizes stable regimes, gradual variations, turning points, coordinated abrupt changes, and potential boundary locations. Only observed features are used; target-well labels are never supplied.

\textbf{Historical classification tool.}
Because intervals are processed from shallow to deep, the history tool can use finalized predictions immediately preceding the current interval:
\begin{equation}
	E_b^{\mathrm{history}}
	=
	\left\{
		\hat{y}_{s_b-h},
		\ldots,
		\hat{y}_{s_b-1}
	\right\},
\end{equation}
where indices are clipped at the beginning of the well, and $h$ is the historical range. The history is empty for the first interval. These predictions are soft context and cannot override a transition supported by the observed logs.

All tool outputs are normalized into the shared record $\mathcal{E}_b$. The same profile is supplied to every view, while role instructions determine which evidence components receive primary attention.
\begin{table}[t]
\centering
\caption{Statistics of each dataset in the experiments.}
\renewcommand{\arraystretch}{1.05}
\label{dataset}
\begin{tabular}{l|cccc}
\hline
Statistics & SEAM & Facies & FORCE & GeoLink \\ \hline
\#of total wells & 5 & 7 & 11 & 128 \\
\#of total samples & 7,092 & 3,164 & 52,766 & 580,205 \\
\#of used input channels & 9 & 7 & 9 & 9 \\
\#of total labels & 7 & 9 & 5 & 11 \\
\#of sampling interval (m) & 10 & 0.5 & 0.15 & 0.125 \\ \hline
\end{tabular}
\end{table}

\subsubsection{Multi-Perspective Reasoning and Fusion}

GeoDecider uses three separate role-conditioned calls that do not observe one another's outputs~\cite{mondorf2024beyond}. The \emph{Data Scientist} emphasizes fast predictions, local measurements, and retrieved analogical cases. The \emph{Context Scientist} focuses on depth-wise trends, candidate boundaries, and preceding finalized predictions. The \emph{Knowledge Scientist} evaluates candidate classes using feature semantics, class descriptions, petrophysical compatibility, and geology-informed guidelines.
Each view returns a candidate sequence for the complete interval together with the identifiers of the evidence supporting its decisions:
\begin{equation}
	\hat{\mathbf{Y}}^{(v)}_b = R_v(\mathcal{E}_b),
	\qquad
	v \in
	\{\mathrm{data}, \mathrm{context}, \mathrm{knowledge}\},
\end{equation}
where $\hat{\mathbf{Y}}^{(v)}_b\in\mathcal{C}^{|\mathcal{B}_b|}$. Their outputs are subsequently reconciled by an evidence-aware fusion process:
\begin{equation}
	\widetilde{\mathbf{Y}}_b
	=
	R_{\mathrm{fuse}}
	\left(
		\{\hat{\mathbf{Y}}^{(v)}_b\}_v,
		\mathcal{E}_b
	\right).
\end{equation}
Unlike deterministic majority voting, $R_{\mathrm{fuse}}$ may retain a majority candidate or select an alternative when the associated evidence more strongly supports it. The fusion output records the competing candidates, cited evidence, and reason for each resolved conflict. This makes disagreement visible rather than hiding it behind an arbitrary tie-breaking rule.
\begin{table*}[htbp]
  \centering
\caption{Classification performance comparison of GeoDecider and other baseline models across four datasets. The best results are highlighted in bold, while the \underline{underlined} results indicate the second-best performance.}
\label{main_results}
 \renewcommand{\arraystretch}{1.05}
  \resizebox{\textwidth}{!}{
  \begin{tabular}{lcccccccccccc}
    \toprule
    \multirow{2}{*}{Method} & \multicolumn{3}{c}{SEAM} & \multicolumn{3}{c}{Facies} & \multicolumn{3}{c}{FORCE} & \multicolumn{3}{c}{GeoLink} \\
    \cmidrule(lr){2-4} \cmidrule(lr){5-7} \cmidrule(lr){8-10} \cmidrule(lr){11-13}
    & Precision & Recall & F1 & Precision & Recall & F1 & Precision & Recall & F1 & Precision & Recall & F1 \\
    \midrule
    XGBoost       & 0.7889 & 0.7188 & 0.7355 & 0.4532 & 0.4388 & 0.4357 & 0.4799 & 0.4597 & 0.4573 & 0.5110 & 0.4185 & 0.4347 \\
    nn-DTW        & 0.6299 & 0.4275 & 0.4396 & 0.4095 & 0.3631 & 0.3589 & 0.3806 & 0.3701 & 0.3616 & 0.3953 & 0.3710 & 0.3675 \\
    GBDT          & 0.8006 & 0.7546 & 0.7626 & \underline{0.4883} & 0.4388 & \underline{0.4440} & 0.4763 & 0.4401 & 0.4445 & 0.4944 & 0.4038 & 0.4149 \\
    LSTMFCN       & 0.8217 & 0.7559 & 0.7694 & 0.4318 & 0.3942 & 0.3869 & 0.4679 & 0.4054 & 0.4046 & \underline{0.5154} & 0.3929 & 0.3983 \\
    MLP           & 0.8462 & 0.7512 & 0.7647 & 0.4532 & 0.4212 & 0.4201 & \underline{0.5935} & \underline{0.5073} & \underline{0.5068} & \textbf{0.5166} & 0.4442 & \underline{0.4501} \\
    MiniRocket    & 0.8457 & 0.7485 & 0.7667 & 0.4615 & 0.4129 & 0.4129 & 0.5288 & 0.4791 & 0.4783 & 0.4857 & 0.4159 & 0.4252 \\
    InceptionTime & 0.8598 & 0.7546 & 0.7667 & 0.4444 & 0.3641 & 0.3817 & 0.4919 & 0.4342 & 0.4450 & 0.4575 & 0.3969 & 0.4083 \\
    InstructTime  & \textbf{0.8684} & 0.7552 & 0.7659 & 0.4371 & 0.3797 & 0.3821 & 0.4537 & 0.4518 & 0.4448 & 0.4583 & 0.4025 & 0.4082 \\
    TableTime     & 0.7912 & 0.6015 & 0.6288 & 0.4296 & 0.3506 & 0.3655 & 0.4719 & 0.3960 & 0.3912 & 0.4061 & 0.3778 & 0.3768 \\
    GPT4TS        & 0.8042 & 0.7330 & 0.7489 & \textbf{0.4911} & \underline{0.4409} & 0.4412 & 0.5118 & 0.4555 & 0.4651 & 0.4805 & 0.4073 & 0.4172 \\
    UniTS         & 0.8005 & \underline{0.7707} & \underline{0.7745} & 0.4471 & 0.4025 & 0.3956 & 0.4418 & 0.4581 & 0.4332 & 0.5077 & \underline{0.4511} & 0.4492 \\
    MOMENT        & 0.7905 & 0.7195 & 0.7315 & 0.4379 & 0.3714 & 0.3887 & 0.4434 & 0.4286 & 0.4270 & 0.4488 & 0.3828 & 0.3961 \\
    \midrule
    \rowcolor[gray]{.9} \textbf{GeoDecider} & \underline{0.8628} & \textbf{0.7748} & \textbf{0.7803} & 0.4758 & \textbf{0.4461} & \textbf{0.4481} & \textbf{0.6094} & \textbf{0.5271} & \textbf{0.5254} & 0.5040 & \textbf{0.4539} & \textbf{0.4582} \\
    \bottomrule
  \end{tabular}
  }
\end{table*}

\subsection{Physics- and Geology-Informed Refinement}
The reconciled sequence may still contain changes that are weakly supported by the observed depth-wise context. GeoDecider therefore applies a physics- and geology-informed refinement process:
\begin{equation}
	\hat{\mathbf{Y}}^{\mathrm{slow}}_b
	=
	R_{\mathrm{refine}}
	\left(
		\widetilde{\mathbf{Y}}_b,
		\mathcal{E}_b
	\right).
\end{equation}

The refinement process checks whether label changes coincide with coordinated variations in surrounding well-log signals, whether isolated predictions are supported by a plausible boundary, and whether the selected class is compatible with observed petrophysical measurements. A revision must identify the evidence that motivates it; otherwise, the fused candidate is retained. These are soft, geology-informed checks rather than a proof of physical validity or a complete geological model.
Combining interval activation with the fast and slow outputs gives, for every $t\in\mathcal{B}_b$,
\begin{equation}
	\hat{y}_t
	=
		\begin{cases}
			\hat{y}^{\mathrm{fast}}_t, & a_b=0,\\
			\hat{y}^{\mathrm{slow}}_{b,t}, & a_b=1.
		\end{cases}
\end{equation}
Thus, sample difficulty is estimated point-wise, whereas an activated interval is interpreted jointly and may revise both anchor and neighboring positions. The final operational sequence is $\hat{\mathbf{Y}}=[\hat{y}_1,\ldots,\hat{y}_L]$, which instantiates the mapping $\mathcal{F}$ defined in the problem formulation. For each activated interval, GeoDecider additionally retains an evidence trace containing the tool records, three candidate sequences, fusion conflicts, and refinement changes.

\section{Experiments}
In this section, we evaluate GeoDecider across multiple datasets against representative baselines, analyze module effectiveness, and discuss practical industrial challenges and lessons learned under privacy and compliance constraints.
\subsection{Experimental Settings}
\subsubsection{Datasets}
We evaluate our method on public well-log datasets from established repositories:
(1) \textbf{SEAM}: the SEG wiki open data catalog for reproducible geoscience benchmarking,
(2) \textbf{Facies}: the Kaggle dataset, where our benchmark split uses seven wells from the Council Grove gas reservoir (Kansas),
(3) \textbf{FORCE}: the Kaggle well-log dataset for lithology prediction, associated with the FORCE 2020 lithology prediction context,
(4) \textbf{GeoLink}: the GEOLINK-S2 well-log dataset with analysis notebooks and preprocessing code.
Details can be found in Table~\ref{dataset} and appendix~\ref{feature_more},~\ref{dataset_more}, and~\ref{app:split_protocol}.

\subsubsection{Baselines}
To conduct a comprehensive and fair comparison, we compare against several baselines grouped into four categories:
\textbf{Machine learning-based} methods: XGBoost~\cite{chen2015xgboost}, nn-DTW~\cite{DTW}, GBDT~\cite{ke2017lightgbm}.
\textbf{Deep learning-based} methods: LSTMFCN~\cite{karim2019multivariate}, MLP~\cite{taud2017multilayer}, MiniRocket~\cite{dempster2021minirocket}, InceptionTime~\cite{ismail2020inceptiontime}.
\textbf{LLM-based} approaches: InstructTime~\cite{cheng2025instructime}, TableTime~\cite{wang2025tabletime}, GPT4TS~\cite{zhou2023one}.
\textbf{Time-series foundation models}: UniTS~\cite{gao2024units} and MOMENT~\cite{goswami2024moment}. For a detailed introduction to these baselines, please see appendix~\ref{Appendix_baseline}.

\subsubsection{Evaluation Metrics} 
To evaluate well-log target-class classification performance, we select three widely used metrics~\cite{zheng2014time,tonutti2019robust}, i.e., Precision, Recall, and F1 measure~\cite{liu2011enhancing,li2021cross}.
Considering our task is a multi-class classification problem, we use the weighted average scores to evaluate the performance of our proposed methods and all baselines. Specifically, we weight the metrics of each class by the number of samples from that class.

\subsubsection{Implementation Details}
In this work, we adopt DeepSeek-R1~\cite{guo2025deepseek} as the base LLM due to its open-source availability and accessible reasoning traces. Each configuration is evaluated over three runs, with predictions aggregated by majority voting, which is distinct from evidence-aware fusion among three scientific views within each run. Inference uses the official API with temperature=0.6, top-p=0.7, and max-tokens=8,192. Deep learning baselines and base classifiers are trained with Adam~\cite{kingma2014adam} in PyTorch on a single NVIDIA GeForce RTX 4090D GPU.

\subsection{Classification Results Analysis}
Table~\ref{main_results} summarizes the classification performance of GeoDecider and representative baselines across four well-log benchmarks. GeoDecider consistently achieves the best performance, obtaining the highest F1 scores on all datasets. Compared with conventional machine learning methods and deep sequence models, GeoDecider demonstrates that geological interpretation benefits from contextual and domain-specific evidence beyond direct feature-to-label mapping.
The improvements are particularly evident on challenging benchmarks such as Facies and FORCE, where logging responses show ambiguity among neighboring lithological units. By activating deliberation on difficult intervals, GeoDecider integrates depth-wise trends, retrieved cases, geological knowledge, and previous predictions to resolve difficult samples. This evidence-grounded process produces more consistent interpretations than fixed prediction.
Furthermore, GeoDecider remains competitive against foundation-model-based approaches, showing that general-purpose sequence models alone may not fully capture geological constraints. 
Overall, the results validate combining efficient prediction with tool-assisted deliberative reasoning for 
reliable geological interpretation.

\subsection{Effectiveness across Initial Predictors}
\begin{figure}[t]
    \centering
    \includegraphics[width=1\linewidth]{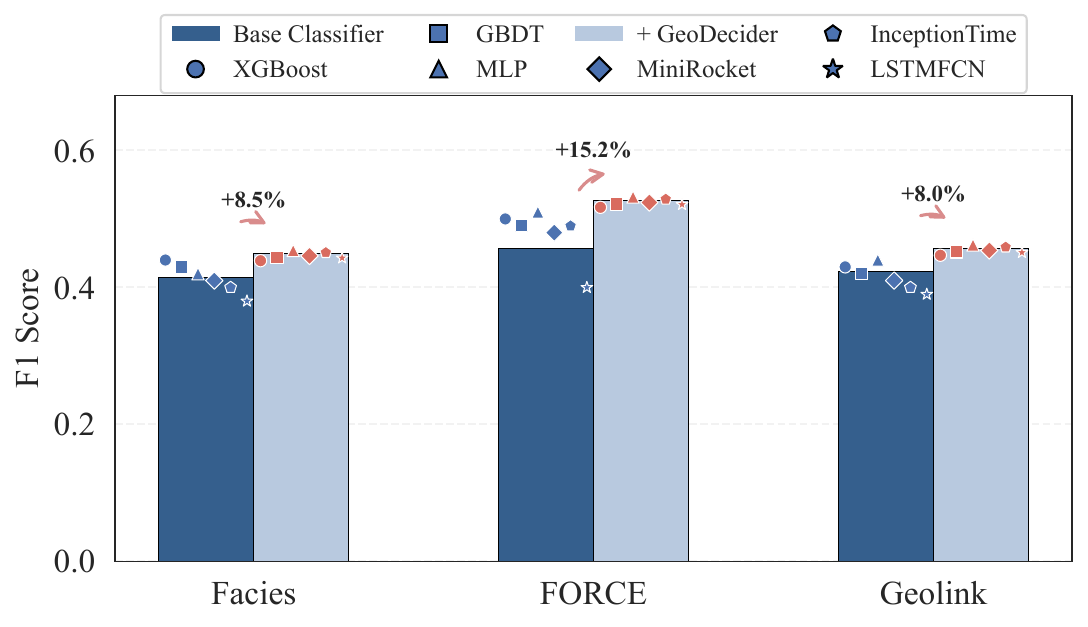}
    \caption{Performance comparison between lightweight models and their refined counterparts, evaluated by F1 score.}
    \label{general}
    \Description{Bar chart comparing base classifiers with their GeoDecider-refined counterparts across datasets.}
\end{figure}

As shown in Figure~\ref{general}, to evaluate the general effectiveness of GeoDecider across initial predictors, we apply it to multiple lightweight classifiers and compare performance before and after refinement. GeoDecider consistently improves F1 scores across datasets, demonstrating classifier-independent effectiveness. The largest gains appear on challenging benchmarks such as FORCE, where geological evidence helps correct ambiguous predictions. These improvements across multiple classifiers indicate that GeoDecider enhances existing predictions through evidence-grounded reasoning rather than stronger representation learning. By integrating multiple evidence sources, GeoDecider refines inconsistent results under geological constraints. The results validate the fast-slow workflow, where efficient models provide initial predictions and deliberation focuses on difficult intervals with improved reliability.
\subsection{Effectiveness of Difficulty-Aware Routing}
\begin{figure}[t]
    \centering
    \includegraphics[width=0.95\linewidth]{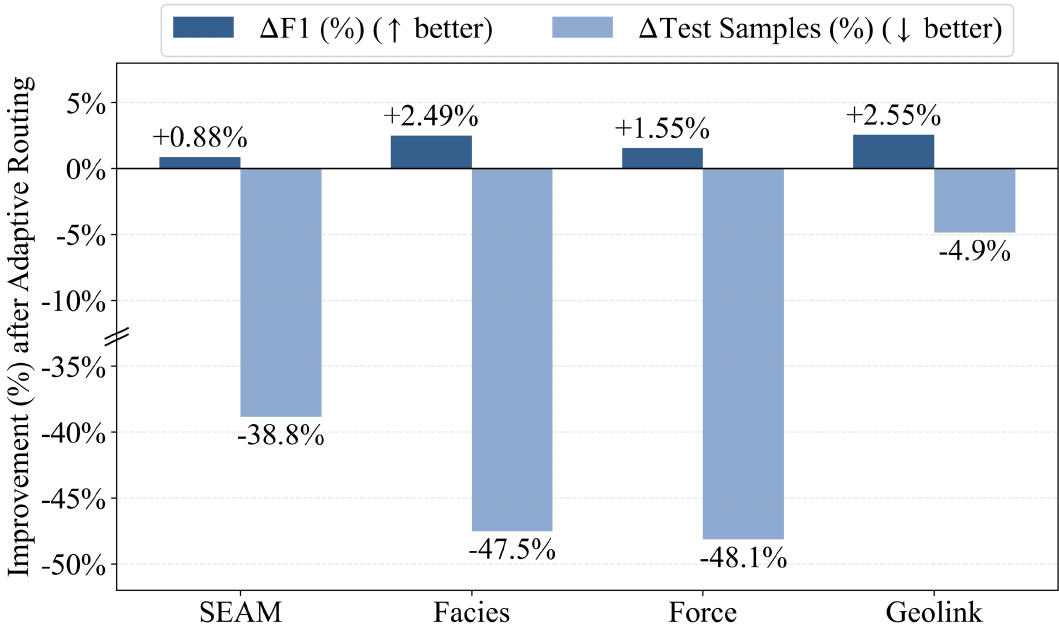}
    \caption{Changes in F1 and point-level routing anchors after difficulty-aware routing. Anchor counts characterize routing selectivity rather than end-to-end LLM cost.}
    \Description{Bar chart showing weighted F1 improvement and the reduction in point-level cases selected as routing anchors across four datasets.}
    \label{yuzhi}
\end{figure}

Figure~\ref{yuzhi} shows that difficulty-aware routing improves weighted F1 across all four datasets while reducing the number of point-level cases selected as routing anchors. We evaluate routing by comparing the full model with a non-selective deliberation variant that activates additional reasoning without difficulty-based filtering. These counts show that prediction-score-based difficulty estimation concentrates additional analysis around a subset of difficult observations.
Importantly, the slow stage operates on complete activated intervals rather than on isolated routing anchors. Neighboring lower-difficulty positions may therefore be included in the same prompt and revised during evidence reconciliation. Accordingly, Figure~\ref{yuzhi} characterizes the selectivity of the routing signal, not end-to-end inference cost; claims about efficiency require direct measurements of model calls.

\subsection{Effectiveness of the Toolkit}

As shown in Table~\ref{ablation}, removing the entire toolkit leads to the largest performance degradation across all benchmarks, confirming that the structured evidence profile is essential for LLM-based lithology reasoning. Among the evaluated components, depth-wise trend analysis contributes the most, as its removal causes the largest single-tool degradation, highlighting the importance of depth-wise continuity and stratigraphic context. The knowledge base improves performance by providing geological priors for ambiguous well-log responses, while neighbor retrieval offers complementary fine-grained analogies. Furthermore, geological refinement enhances consistency by validating predictions against local transitions. These results confirm that different evidence sources play complementary roles in grounded lithology reasoning.
\begin{table}[t]
\centering
\caption{Comprehensive ablation on the impact of the toolkit and geological refinement module, evaluated by F1 score.}
\label{ablation}
\renewcommand{\arraystretch}{1.05}
\resizebox{\linewidth}{!}{%
\begin{tabular}{lcccc}
\toprule
Variant & SEAM & Facies & FORCE & GeoLink \\ \midrule
\cellcolor{lightgray!30}\textbf{Full Model} & \cellcolor{lightgray!30}\textbf{0.7803} & \cellcolor{lightgray!30}\textbf{0.4481} & \cellcolor{lightgray!30}\textbf{0.5254} & \cellcolor{lightgray!30}\textbf{0.4582} \\ \midrule
$w/o$ Toolkit & 0.7642 & 0.4312 & 0.4985 & 0.4452 \\
$w/o$ Knowledge Base & 0.7745 & 0.4405 & 0.5132 & 0.4534 \\
$w/o$ Depth-Wise Trend Analysis & 0.7702 & 0.4362 & 0.5075 & 0.4505 \\
$w/o$ Neighbor Retrieval & \underline{0.7791} & \underline{0.4452} & \underline{0.5198} & \underline{0.4561} \\
$w/o$ Geological Refinement & 0.7742 & 0.4402 & 0.5135 & 0.4536 \\
\bottomrule
\end{tabular}%
}
\end{table}


\begin{figure*}[t]
	\centering
\includegraphics[width=1\textwidth]{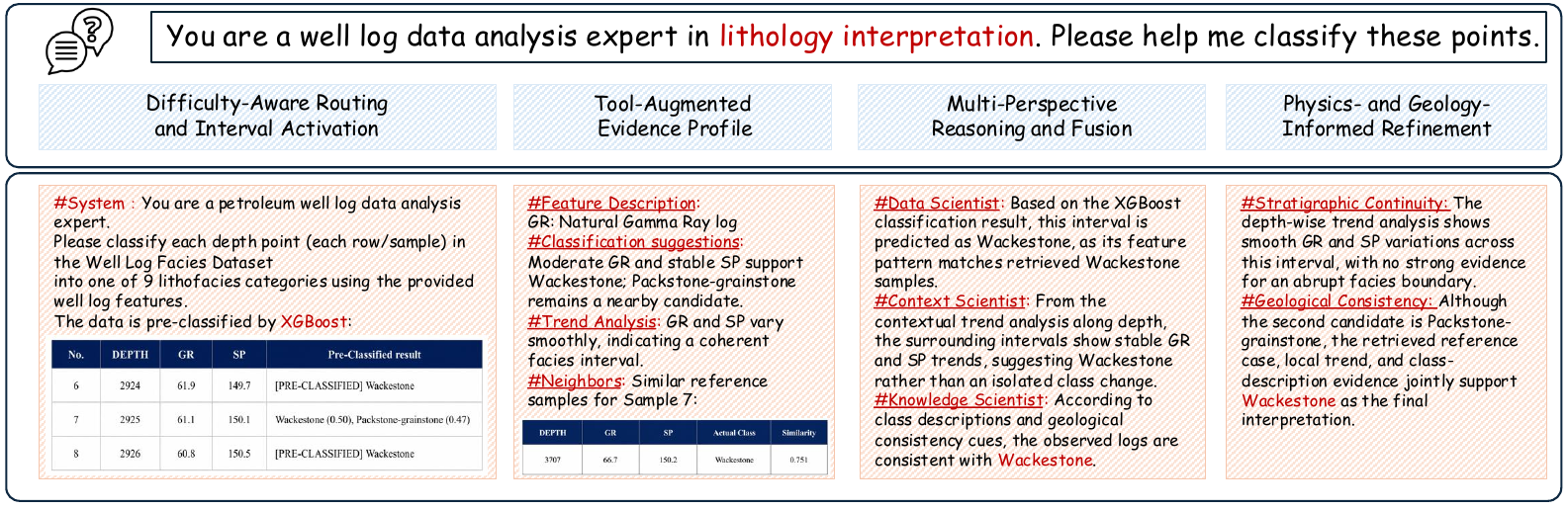}
	\caption{A case of how GeoDecider performs well-log target-class classification via pre-classified and tool-augmented reasoning.}
    \Description{Case study showing base predictions, tool-augmented reasoning outputs, and refined target-class predictions.}
	\label{case}
\end{figure*}

\subsection{In-Depth Analysis of Geological Refinement}
Table~\ref{suipianhua} shows that geological refinement consistently reduces the Flying Point Ratio (FPR) across all benchmarks. To evaluate local sequence continuity after refinement, FPR measures isolated label predictions along depth. Given $\hat{\mathbf{Y}}=[\hat{y}_1,\ldots,\hat{y}_L]$, a flying point is a prediction differing from all labels within its local neighborhood. A lower FPR indicates stronger depth-wise continuity. The reduction demonstrates the ability of refinement to suppress spurious point-wise predictions and correct locally inconsistent labels using contextual evidence. However, FPR only measures local continuity and does not represent complete geological validity.

\begin{figure}[t]
    \centering
    \includegraphics[width=1\linewidth]{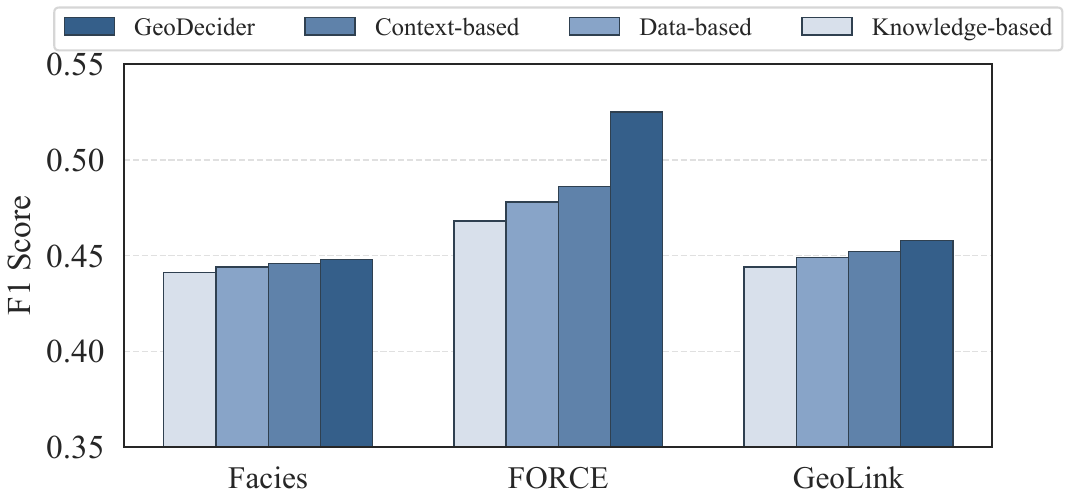}
    \caption{Performance comparison between single-view reasoning and integrated multi-view reasoning across benchmarks for lithology interpretation. }
    \label{tuili}
    \Description{Bar chart comparing single-perspective reasoning variants with multi-perspective GeoDecider reasoning.}
\end{figure}

\subsection{Analysis of Multi-Perspective Reasoning}
Figure~\ref{tuili} shows that GeoDecider consistently achieves the best weighted F1 across all datasets. This experiment evaluates multi-perspective reasoning by comparing GeoDecider with three single-perspective variants based on the context, data, and knowledge views. The gains are particularly evident on FORCE, where combining perspectives outperforms individual strategies. Different perspectives provide complementary information: the context view captures stratigraphic continuity, the data view models local patterns, and the knowledge view enforces geological constraints. 

\subsection{Case Study}
Figure~\ref{case} presents a case study illustrating GeoDecider's end-to-end decision process for well-log target-class classification. The base classifier provides point-wise pre-classification based on log features. High-difficulty routing anchors activate interval-level tool-augmented reasoning that incorporates feature descriptions, trend analysis, historical predictions, and neighboring reference cases. These signals are examined from multiple perspectives and reconciled according to their cited evidence. The refinement module then checks the fused sequence against observed geological patterns and local continuity cues to produce a contextually supported final prediction. This case demonstrates how GeoDecider combines evidence to generate coherent, evidence-supported and reliable target-class interpretations.
\begin{table}[t]
\centering
\caption{Flying Point Ratio (FPR) of the base classifier and GeoDecider with and without geological refinement.}
\label{suipianhua}
\resizebox{\columnwidth}{!}{%
\begin{tabular}{lcccc}
\toprule
Variant & SEAM & Facies & FORCE & GeoLink \\
\midrule
Base Classifier & 0.0148 & 0.1760 & 0.0873 & 0.1240 \\
GeoDecider(w/o Ref) & \underline{0.0141} & \underline{0.1596} & \underline{0.0792} & \underline{0.1196} \\
GeoDecider(w/ Ref) & \textbf{0.0135} & \textbf{0.1541} & \textbf{0.0746} & \textbf{0.1137} \\
\bottomrule
\end{tabular}%
}
\end{table}

\subsection{Limitations and Ethical Considerations}

All experiments use public well-log benchmarks. GeoDecider was developed with a national oil company using proprietary data, but confidentiality prevents disclosure of raw logs, locations, and production results. Industrial collaboration only informs deployment constraints and design. Proprietary applications should follow data-governance policies and use de-identified results. 

\subsection{Generative AI Usage}
The method applies large language models to difficult intervals using structured Evidence Profiles and role-conditioned prompts. No task-specific fine-tuning is performed. Authors designed prompts, verified outputs, and reviewed all claims, experiments, results, citations, and manuscript content, taking full responsibility.

\section{Conclusion}

In this work, we presented GeoDecider, a scientific agent for geological interpretation via evidence-grounded deliberation. GeoDecider transforms single-pass well-log classification into an adaptive fast-slow workflow, where lightweight models handle routine predictions and difficult intervals trigger evidence-based reasoning. Specialized tools construct Evidence Profiles from geological knowledge, depth-wise trends, analogical cases, and historical predictions. GeoDecider integrates complementary perspectives, reconciles conflicting evidence, and applies geology-informed refinement. Experiments on four public benchmarks demonstrate consistent improvements over representative baselines, highlighting the value of geological context for ambiguous intervals. Overall, GeoDecider shows that reliable interpretation requires not only accurate prediction, but also structured evidence integration, multi-perspective reasoning, and transparent decision-making.

\clearpage

\bibliographystyle{ACM-Reference-Format}
\balance
\bibliography{ref}

@inproceedings{zheng2014time,
  title={Time series classification using multi-channels deep convolutional neural networks},
  author={Zheng, Yi and Liu, Qi and Chen, Enhong and Ge, Yong and Zhao, J Leon},
  booktitle={International conference on web-age information management},
  pages={298--310},
  year={2014},
  organization={Springer}
}

@article{tonutti2019robust,
  title={Robust and subject-independent driving manoeuvre anticipation through domain-adversarial recurrent neural networks},
  author={Tonutti, Michele and Ruffaldi, Emanuele and Cattaneo, Alessandro and Avizzano, Carlo Alberto},
  journal={Robotics and Autonomous Systems},
  volume={115},
  pages={162--173},
  year={2019},
  publisher={Elsevier}
}

@article{liu2011enhancing,
  title={Enhancing collaborative filtering by user interest expansion via personalized ranking},
  author={Liu, Qi and Chen, Enhong and Xiong, Hui and Ding, Chris HQ and Chen, Jian},
  journal={IEEE Transactions on Systems, Man, and Cybernetics, Part B (Cybernetics)},
  volume={42},
  number={1},
  pages={218--233},
  year={2011},
  publisher={IEEE}
}

@article{karpatne2018machine,
  title={Machine learning for the geosciences: Challenges and opportunities},
  author={Karpatne, Anuj and Ebert-Uphoff, Imme and Ravela, Sai and Babaie, Hassan Ali and Kumar, Vipin},
  journal={IEEE Transactions on Knowledge and Data Engineering},
  volume={31},
  number={8},
  pages={1544--1554},
  year={2018},
  publisher={IEEE}
}

@article{bergen2019machine,
  title={Machine learning for data-driven discovery in solid Earth geoscience},
  author={Bergen, Karianne J and Johnson, Paul A and de Hoop, Maarten V and Beroza, Gregory C},
  journal={Science},
  volume={363},
  number={6433},
  pages={eaau0323},
  year={2019},
  publisher={American Association for the Advancement of Science}
}

@article{hall2016facies,
  title={Facies classification using machine learning},
  author={Hall, Brendon},
  journal={The Leading Edge},
  volume={35},
  number={10},
  pages={906--909},
  year={2016},
  publisher={Society of Exploration Geophysicists}
}

@article{DTW,
  title={Using dynamic time warping distances as features for improved time series classification},
  author={Kate, Rohit J},
  journal={Data mining and knowledge discovery},
  volume={30},
  pages={283--312},
  year={2016},
  publisher={Springer}
}

@article{InceptionTime,
  title={Inceptiontime: Finding alexnet for time series classification},
  author={Ismail Fawaz, Hassan and Lucas, Benjamin and Forestier, Germain and Pelletier, Charlotte and Schmidt, Daniel F and Weber, Jonathan and Webb, Geoffrey I and Idoumghar, Lhassane and Muller, Pierre-Alain and Petitjean, Fran{\c{c}}ois},
  journal={Data Mining and Knowledge Discovery},
  volume={34},
  number={6},
  pages={1936--1962},
  year={2020},
  publisher={Springer}
}

@article{chen2015xgboost,
  title={Xgboost: extreme gradient boosting},
  author={Chen, Tianqi and He, Tong and Benesty, Michael and Khotilovich, Vadim and Tang, Yuan and Cho, Hyunsu and Chen, Kailong and Mitchell, Rory and Cano, Ignacio and Zhou, Tianyi and others},
  journal={R package version 0.4-2},
  volume={1},
  number={4},
  pages={1--4},
  year={2015}
}

@inproceedings{ke2017lightgbm,
author = {Ke, Guolin and Meng, Qi and Finley, Thomas and Wang, Taifeng and Chen, Wei and Ma, Weidong and Ye, Qiwei and Liu, Tie-Yan},
title = {LightGBM: a highly efficient gradient boosting decision tree},
year = {2017},
isbn = {9781510860964},
publisher = {Curran Associates Inc.},
address = {Red Hook, NY, USA},
booktitle = {Proceedings of the 31st International Conference on Neural Information Processing Systems},
pages = {3149–3157},
numpages = {9},
location = {Long Beach, California, USA},
series = {NIPS'17}
}

@article{karim2019multivariate,
  title={Multivariate LSTM-FCNs for time series classification},
  author={Karim, Fazle and Majumdar, Somshubra and Darabi, Houshang and Harford, Samuel},
  journal={Neural networks},
  volume={116},
  pages={237--245},
  year={2019},
  publisher={Elsevier}
}

@incollection{taud2017multilayer,
  title={Multilayer perceptron (MLP)},
  author={Taud, Hind and Mas, Jean-Franccois},
  booktitle={Geomatic approaches for modeling land change scenarios},
  pages={451--455},
  year={2017},
  publisher={Springer}
}

@inproceedings{dempster2021minirocket,
author = {Dempster, Angus and Schmidt, Daniel F. and Webb, Geoffrey I.},
title = {MiniRocket: A Very Fast (Almost) Deterministic Transform for Time Series Classification},
year = {2021},
isbn = {9781450383325},
publisher = {Association for Computing Machinery},
address = {New York, NY, USA},
url = {https://doi.org/10.1145/3447548.3467231},
doi = {10.1145/3447548.3467231},
booktitle = {Proceedings of the 27th ACM SIGKDD Conference on Knowledge Discovery \& Data Mining},
pages = {248–257},
numpages = {10},
location = {Virtual Event, Singapore},
series = {KDD '21}
}

@article{ismail2020inceptiontime,
  title={Inceptiontime: Finding alexnet for time series classification},
  author={Ismail Fawaz, Hassan and Lucas, Benjamin and Forestier, Germain and Pelletier, Charlotte and Schmidt, Daniel F and Weber, Jonathan and Webb, Geoffrey I and Idoumghar, Lhassane and Muller, Pierre-Alain and Petitjean, Fran{\c{c}}ois},
  journal={Data Mining and Knowledge Discovery},
  volume={34},
  number={6},
  pages={1936--1962},
  year={2020},
  publisher={Springer}
}

@inproceedings{wang2025tabletime,
  title={Tabletime: Reformulating time series classification as training-free table understanding with large language models},
  author={Wang, Jiahao and Cheng, Mingyue and Mao, Qingyang and Zhou, Yitong and Wang, Daoyu and Liu, Qi and Xu, Feiyang and Li, Xin},
  booktitle={Proceedings of the 34th ACM International Conference on Information and Knowledge Management},
  pages={3009--3019},
  year={2025}
}

@article{gao2024units,
  title={Units: A unified multi-task time series model},
  author={Gao, Shanghua and Koker, Teddy and Queen, Owen and Hartvigsen, Tom and Tsiligkaridis, Theodoros and Zitnik, Marinka},
  journal={Advances in Neural Information Processing Systems},
  volume={37},
  pages={140589--140631},
  year={2024}
}

@inproceedings{goswami2024moment,
author = {Goswami, Mononito and Szafer, Konrad and Choudhry, Arjun and Cai, Yifu and Li, Shuo and Dubrawski, Artur},
title = {MOMENT: a family of open time-series foundation models},
year = {2024},
publisher = {JMLR.org},
booktitle = {Proceedings of the 41st International Conference on Machine Learning},
articleno = {642},
numpages = {38},
location = {Vienna, Austria},
series = {ICML'24}
}

@inproceedings{li2021cross,
  title={Cross-oilfield reservoir classification via multi-scale sensor knowledge transfer},
  author={Li, Zhi and Wang, Zhefeng and Wei, Zhicheng and Zhou, Xiangguang and Wang, Yijun and Huai, Baoxing and Liu, Qi and Yuan, Nicholas Jing and Gong, Renbin and Chen, Enhong},
  booktitle={Proceedings of the AAAI Conference on Artificial Intelligence},
  volume={35},
  number={5},
  pages={4215--4223},
  year={2021}
}

@inproceedings{zhang2024large,
author = {Zhang, Xiyuan and Chowdhury, Ranak Roy and Gupta, Rajesh K. and Shang, Jingbo},
title = {Large language models for time series: a survey},
year = {2024},
isbn = {978-1-956792-04-1},
url = {https://doi.org/10.24963/ijcai.2024/921},
doi = {10.24963/ijcai.2024/921},
booktitle = {Proceedings of the Thirty-Third International Joint Conference on Artificial Intelligence},
articleno = {921},
numpages = {9},
location = {Jeju, Korea},
series = {IJCAI '24}
}

@inproceedings{jiang2024empowering,
author = {Jiang, Yushan and Pan, Zijie and Zhang, Xikun and Garg, Sahil and Schneider, Anderson and Nevmyvaka, Yuriy and Song, Dongjin},
title = {Empowering time series analysis with large language models: a survey},
year = {2024},
isbn = {978-1-956792-04-1},
url = {https://doi.org/10.24963/ijcai.2024/895},
doi = {10.24963/ijcai.2024/895},
booktitle = {Proceedings of the Thirty-Third International Joint Conference on Artificial Intelligence},
articleno = {895},
numpages = {9},
location = {Jeju, Korea},
series = {IJCAI '24}
}

@inproceedings{jin2024position,
author = {Jin, Ming and Zhang, Yifan and Chen, Wei and Zhang, Kexin and Liang, Yuxuan and Yang, Bin and Wang, Jindong and Pan, Shirui and Wen, Qingsong},
title = {Position: what can large language models tell us about time series analysis},
year = {2024},
publisher = {JMLR.org},
booktitle = {Proceedings of the 41st International Conference on Machine Learning},
articleno = {895},
numpages = {17},
location = {Vienna, Austria},
series = {ICML'24}
}

@article{chang2023llm4ts,
  title={Llm4ts: Two-stage fine-tuning for time-series forecasting with pre-trained llms},
  author={Chang, Ching and Peng, Wen-Chih and Chen, Tien-Fu},
  journal={CoRR},
  year={2023}
}

@article{zhou2023one,
  title={One fits all: Power general time series analysis by pretrained lm},
  author={Zhou, Tian and Niu, Peisong and Sun, Liang and Jin, Rong and others},
  journal={Advances in neural information processing systems},
  volume={36},
  pages={43322--43355},
  year={2023}
}

@article{nakano2021webgpt,
  title={Webgpt: Browser-assisted question-answering with human feedback},
  author={Nakano, Reiichiro and Hilton, Jacob and Balaji, Suchir and Wu, Jeff and Ouyang, Long and Kim, Christina and Hesse, Christopher and Jain, Shantanu and Kosaraju, Vineet and Saunders, William and others},
  journal={arXiv preprint arXiv:2112.09332},
  year={2021}
}

@article{tan2024language,
  title={Are language models actually useful for time series forecasting?},
  author={Tan, Mingtian and Merrill, Mike and Gupta, Vinayak and Althoff, Tim and Hartvigsen, Tom},
  journal={Advances in Neural Information Processing Systems},
  volume={37},
  pages={60162--60191},
  year={2024}
}

@inproceedings{NEURIPS2024_053ee34c,
author = {Liu, Jingwei and Yang, Ling and Li, Hongyan and Hong, Shenda},
title = {Retrieval-augmented diffusion models for time series forecasting},
year = {2024},
isbn = {9798331314385},
publisher = {Curran Associates Inc.},
address = {Red Hook, NY, USA},
booktitle = {Proceedings of the 38th International Conference on Neural Information Processing Systems},
articleno = {91},
numpages = {21},
location = {Vancouver, BC, Canada},
series = {NIPS '24}
}

@inproceedings{cheng2025instructime,
author = {Cheng, Mingyue and Chen, Yiheng and Liu, Qi and Liu, Zhiding and Luo, Yucong and Chen, Enhong},
title = {InstrucTime: Advancing Time Series Classification with Multimodal Language Modeling},
year = {2025},
isbn = {9798400713293},
publisher = {Association for Computing Machinery},
address = {New York, NY, USA},
url = {https://doi.org/10.1145/3701551.3703499},
doi = {10.1145/3701551.3703499},
booktitle = {Proceedings of the Eighteenth ACM International Conference on Web Search and Data Mining},
pages = {792–800},
numpages = {9},
location = {Hannover, Germany},
series = {WSDM '25}
}

@article{mondorf2024beyond,
  title={Beyond accuracy: evaluating the reasoning behavior of large Language models--A survey},
  author={Mondorf, Philipp and Plank, Barbara},
  journal={arXiv preprint arXiv:2404.01869},
  year={2024}
}

@article{kingma2014adam,
  title={Adam: A method for stochastic optimization},
  author={Kingma, Diederik P},
  journal={arXiv preprint arXiv:1412.6980},
  year={2014}
}

@article{beloborodov2024automated,
  title={Automated lithofluid and facies classification in well logs: The rock-physics perspective},
  author={Beloborodov, Roman and Gunning, James and Pervukhina, Marina and Hauser, Juerg and Clennell, Michael Ben and Mur, Alan and Li, Vladimir},
  journal={Geophysics},
  volume={89},
  number={4},
  pages={MR209--MR222},
  year={2024},
  publisher={Society of Exploration Geophysicists}
}

@article{dong2023deep,
  title={A deep kernel method for lithofacies identification using conventional well logs},
  author={Dong, Shao-Qun and Zhong, Zhao-Hui and Cui, Xue-Hui and Zeng, Lian-Bo and Yang, Xu and Liu, Jian-Jun and Sun, Yan-Ming and Hao, Jing-Ru},
  journal={Petroleum Science},
  volume={20},
  number={3},
  pages={1411--1428},
  year={2023},
  publisher={Elsevier}
}

@article{li2025large,
  title={A large-scale, high-quality dataset for lithology identification: Construction and applications},
  author={Li, Jia-Yu and Tang, Ji-Zhou and Zhao, Xian-Zheng and Fan, Bo and Jiang, Wen-Ya and Song, Shun-Yao and Li, Jian-Bing and Chen, Kai-Da and Zhao, Zheng-Guang},
  journal={Petroleum Science},
  year={2025},
  publisher={Elsevier}
}

@article{wu1987automated,
  title={Automated stratigraphic interpretation of well-log data},
  author={Wu, Xuanzhi and Nyland, Edo},
  journal={Geophysics},
  volume={52},
  number={12},
  pages={1665--1676},
  year={1987},
  publisher={Society of Exploration Geophysicists}
}

@article{shi2023refined,
  title={Refined lithology identification: Methodology, challenges and prospects},
  author={Shi, Heng and Xu, ZhenHao and Lin, Peng and Ma, Wen},
  journal={Geoenergy Science and Engineering},
  volume={231},
  pages={212382},
  year={2023},
  publisher={Elsevier}
}

@article{bressan2020evaluation,
  title={Evaluation of machine learning methods for lithology classification using geophysical data},
  author={Bressan, Thiago Santi and de Souza, Marcelo Kehl and Girelli, Tiago J and Junior, Farid Chemale},
  journal={Computers \& Geosciences},
  volume={139},
  pages={104475},
  year={2020},
  publisher={Elsevier}
}

@article{xu2024geopredict,
  title={GeoPredict-LLM: Intelligent tunnel advanced geological prediction by reprogramming large language models},
  author={Xu, Zhenhao and Wang, Zhaoyang and Li, Shucai and Zhang, Xiao and Lin, Peng},
  journal={Intelligent Geoengineering},
  volume={1},
  number={1},
  pages={49--57},
  year={2024},
  publisher={Elsevier}
}

@article{liu2025logging,
  title={Logging-data-driven lithology identification of conglomerate reservoir by the assistance of integrated machine learning methods},
  author={Liu, Jiming and Xu, Dongjin},
  journal={Scientific Reports},
  year={2025},
  publisher={Nature Publishing Group UK London}
}

@article{saleem2025multiple,
  title={Multiple machine learning algorithms for lithofacies prediction in the deltaic depositional system of the lower Goru Formation, Lower Indus Basin, Pakistan},
  author={Saleem, Muhammad Ansar and Sohail, Ghulam Mohyuddin and Rehman, Saif Ur and Yasin, Qamar and Radwan, Ahmed E},
  journal={Scientific Reports},
  volume={15},
  number={1},
  pages={34933},
  year={2025},
  publisher={Nature Publishing Group UK London}
}

@article{park2022data,
  title={Data-driven sequence labeling methods incorporating the long-range spatial variation of geological data for lithofacies sequence estimation},
  author={Park, Gyeong-Tae and Jeong, Jina and Emelyanova, Irina and Pervukhina, Marina and Esteban, Lionel and Yun, Seong-Taek},
  journal={Journal of Petroleum Science and Engineering},
  volume={208},
  pages={109345},
  year={2022},
  publisher={Elsevier}
}

@article{liu2021lithological,
  title={A lithological sequence classification method with well log via SVM-assisted bi-directional GRU-CRF neural network},
  author={Liu, Zhege and Cao, Junxing and You, Jiachun and Chen, Shuna and Lu, Yujia and Zhou, Peng},
  journal={Journal of Petroleum Science and Engineering},
  volume={205},
  pages={108913},
  year={2021},
  publisher={Elsevier}
}

@article{ruiyi2021lithology,
  title={Lithology identification of igneous rocks based on XGboost and conventional logging curves, a case study of the eastern depression of Liaohe Basin},
  author={Ruiyi, HAN and Zhuwen, WANG and Wenhua, WANG and Fanghui, XU and Xinghua, QI and Yitong, CUI},
  journal={Journal of Applied Geophysics},
  volume={195},
  pages={104480},
  year={2021},
  publisher={Elsevier}
}

@article{rathore2023well,
  title={Well log analysis and comparison of supervised machine learning algorithms for lithofacies identification in pab formation, lower indus basin},
  author={Rathore, Pal Washa Shahzad and Hussain, Matloob and Malik, Muhammad Bilal and Amin, Yawar},
  journal={Journal of Applied Geophysics},
  volume={219},
  pages={105199},
  year={2023},
  publisher={Elsevier}
}

@article{nguyen2025classifying,
  title={Classifying reservoir facies using attention-based residual neural networks},
  author={Nguyen, An Hai and Nguyen, Khang and Mai, Nga},
  journal={PeerJ Computer Science},
  volume={11},
  pages={e2977},
  year={2025},
  publisher={PeerJ Inc.}
}

@inproceedings{kusuma2025leveraging,
  title={Leveraging Hidden Markov Model with Window Filtering Technique for Unsupervised Facies Detection in Well Logs},
  author={Kusuma, Lundi and Djamaoeddin, Alwin},
  booktitle={SPE Middle East Oil and Gas Show and Conference},
  pages={D031S085R006},
  year={2025},
  organization={SPE}
}

@article{schumann2002hidden,
  title={Hidden Markov models for lithological well log classification},
  author={Schumann, Agnes},
  journal={Terra Nostra},
  volume={4},
  pages={373--378},
  year={2002}
}

@article{chen2025geological,
  title={Geological information-driven deep learning for lithology identification from well logs},
  author={Chen, Luoyuan and Wang, Xingjian and Liu, Zhanbo},
  journal={Frontiers in Earth Science},
  volume={13},
  pages={1662760},
  year={2025},
  publisher={Frontiers Media SA}
}

@article{jiang2024machine,
  title={Machine learning (ML) for fluvial lithofacies identification from well logs: A hybrid classification model integrating lithofacies characteristics, logging data distributions, and ML models applicability},
  author={Jiang, Shiyi and Sun, Panke and Lyu, Fengqing and Zhu, Sicheng and Zhou, Ruifeng and Li, Bin and He, Taihong and Lin, Yujian and Gao, Yining and Song, Wendan and others},
  journal={Geoenergy Science and Engineering},
  volume={233},
  pages={212587},
  year={2024},
  publisher={Elsevier}
}

@article{fan2025logging,
  title={Logging-data-driven lithology identification in complex reservoirs: an example from the Niuxintuo block of the Liaohe oilfield},
  author={Fan, Zuochun and Hu, Changhao and Jiang, Shu and Li, Man and Cai, Ye and Jiang, Yue and Li, Yang and Tian, Mei},
  journal={Frontiers in Earth Science},
  volume={13},
  pages={1491334},
  year={2025},
  publisher={Frontiers Media SA}
}

@article{pang2025empowering,
  title={Empowering lithology identification with FreLog: Leveraging frequency domain insights in well logging signal processing},
  author={Pang, Qingwei and Chen, Chenglizhao and Sun, Youzhuang and Pang, Shanchen},
  journal={Measurement},
  volume={246},
  pages={116710},
  year={2025},
  publisher={Elsevier}
}

@article{guo2025deepseek,
  title={Deepseek-r1: Incentivizing reasoning capability in llms via reinforcement learning},
  author={Guo, Daya and Yang, Dejian and Zhang, Haowei and Song, Junxiao and Zhang, Ruoyu and Xu, Runxin and Zhu, Qihao and Ma, Shirong and Wang, Peiyi and Bi, Xiao and others},
  journal={arXiv preprint arXiv:2501.12948},
  year={2025}
}

@article{wei2022chain,
  title={Chain-of-thought prompting elicits reasoning in large language models},
  author={Wei, Jason and Wang, Xuezhi and Schuurmans, Dale and Bosma, Maarten and Xia, Fei and Chi, Ed and Le, Quoc V and Zhou, Denny and others},
  journal={Advances in neural information processing systems},
  volume={35},
  pages={24824--24837},
  year={2022}
}

@article{wang2022self,
  title={Self-consistency improves chain of thought reasoning in language models},
  author={Wang, Xuezhi and Wei, Jason and Schuurmans, Dale and Le, Quoc and Chi, Ed and Narang, Sharan and Chowdhery, Aakanksha and Zhou, Denny},
  journal={arXiv preprint arXiv:2203.11171},
  year={2022}
}

@inproceedings{yao2023react,
  title={React: Synergizing reasoning and acting in language models},
  author={Yao, Shunyu and Zhao, Jeffrey and Yu, Dian and Du, Nan and Shafran, Izhak and Narasimhan, Karthik and Cao, Yuan},
  booktitle={International Conference on Learning Representations (ICLR)},
  year={2023}
}

@article{schick2023toolformer,
  title={Toolformer: Language models can teach themselves to use tools},
  author={Schick, Timo and Dwivedi-Yu, Jane and Dess{\`\i}, Roberto and Raileanu, Roberta and Lomeli, Maria and Hambro, Eric and Zettlemoyer, Luke and Cancedda, Nicola and Scialom, Thomas},
  journal={Advances in Neural Information Processing Systems},
  volume={36},
  pages={68539--68551},
  year={2023}
}

@article{wang2023describe,
  title={Describe, explain, plan and select: interactive planning with llms enables open-world multi-task agents},
  author={Wang, Zihao and Cai, Shaofei and Chen, Guanzhou and Liu, Anji and Ma, Xiaojian Shawn and Liang, Yitao},
  journal={Advances in Neural Information Processing Systems},
  volume={36},
  pages={34153--34189},
  year={2023}
}

@article{luo2025time,
  title={Time series forecasting as reasoning: A slow-thinking approach with reinforced llms},
  author={Luo, Yucong and Zhou, Yitong and Cheng, Mingyue and Wang, Jiahao and Wang, Daoyu and Pan, Tingyue and Zhang, Jintao},
  journal={arXiv preprint arXiv:2506.10630},
  year={2025}
}

@inproceedings{liang2024foundation,
  title={Foundation models for time series analysis: A tutorial and survey},
  author={Liang, Yuxuan and Wen, Haomin and Nie, Yuqi and Jiang, Yushan and Jin, Ming and Song, Dongjin and Pan, Shirui and Wen, Qingsong},
  booktitle={Proceedings of the 30th ACM SIGKDD conference on knowledge discovery and data mining},
  pages={6555--6565},
  year={2024}
}

@inproceedings{zhang2025unleashing,
  title={Unleashing the power of pre-trained language models for irregularly sampled time series},
  author={Zhang, Weijia and Yin, Chenlong and Liu, Hao and Xiong, Hui},
  booktitle={Proceedings of the 31st ACM SIGKDD Conference on Knowledge Discovery and Data Mining V. 2},
  pages={3831--3842},
  year={2025}
}

@inproceedings{mohammadi2025evaluation,
  title={Evaluation and benchmarking of llm agents: A survey},
  author={Mohammadi, Mahmoud and Li, Yipeng and Lo, Jane and Yip, Wendy},
  booktitle={Proceedings of the 31st ACM SIGKDD Conference on Knowledge Discovery and Data Mining V. 2},
  pages={6129--6139},
  year={2025}
}

@article{schmidt2009distilling,
  title={Distilling free-form natural laws from experimental data},
  author={Schmidt, Michael and Lipson, Hod},
  journal={science},
  volume={324},
  number={5923},
  pages={81--85},
  year={2009},
  publisher={American Association for the Advancement of Science}
}

@article{gil2014amplify,
  title={Amplify scientific discovery with artificial intelligence},
  author={Gil, Yolanda and Greaves, Mark and Hendler, James and Hirsh, Haym},
  journal={Science},
  volume={346},
  number={6206},
  pages={171--172},
  year={2014},
  publisher={American Association for the Advancement of Science}
}

@article{jumper2021highly,
  title={Highly accurate protein structure prediction with AlphaFold},
  author={Jumper, John and Evans, Richard and Pritzel, Alexander and Green, Tim and Figurnov, Michael and Ronneberger, Olaf and Tunyasuvunakool, Kathryn and Bates, Russ and {\v{Z}}{\'\i}dek, Augustin and Potapenko, Anna and others},
  journal={nature},
  volume={596},
  number={7873},
  pages={583--589},
  year={2021},
  publisher={Nature Publishing Group UK London}
}

@article{reichstein2019deep,
  title={Deep learning and process understanding for data-driven Earth system science},
  author={Reichstein, Markus and Camps-Valls, Gustau and Stevens, Bjorn and Jung, Martin and Denzler, Joachim and Carvalhais, Nuno and Prabhat, F},
  journal={Nature},
  volume={566},
  number={7743},
  pages={195--204},
  year={2019},
  publisher={Nature Publishing Group UK London}
}
\newpage

\appendix

\section*{Appendix}
\section{Dataset Specifications and Input Features}
\label{feature_more}
To facilitate reproducibility and clarify the inputs used by the GeoDecider framework, we provide the detailed configuration for each benchmark dataset. Table \ref{tab:dataset_features} summarizes the specific geophysical logs and features selected as inputs for the model, along with the target interpretation classes.

For the \textbf{Facies} dataset, the 9 classes correspond to specific depositional environments ranging from \textit{Nonmarine sandstone} to \textit{Phylloid-algal bafflestone}. The \textbf{FORCE} dataset targets 5 primary lithologies: \textit{Shale, Sandstone, Limestone, Marl,} and \textit{Sandstone/Shale}. \textbf{GeoLink} includes 11 granular classes, distinguishing between variations such as \textit{Silty Sand}, \textit{Cross Bedded Sand}, and \textit{Argillaceous Limestone}. Finally, \textbf{SEAM} classifies geological ages and salt bodies, including \textit{Mother Salt}, \textit{Cretaceous}, and \textit{Lower Miocene} layers.

\section{Datasets Description}
\label{dataset_more}
We conduct experiments on four publicly available well-log benchmarks to assess the generalization ability of GeoDecider across different geological environments. These datasets provide depth-wise petrophysical measurements and lithology-related annotations, while varying in reservoir characteristics, sampling intervals, and well availability. All experiments follow the original benchmark settings, and no target-well labels are used during inference.
\begin{itemize}
\item \textbf{SEAM}: the SEG Wiki Open Data ``Well logs'' catalog\footnote{\url{https://wiki.seg.org/wiki/Open_data\#Well_logs}} as an entry point to publicly released well-log resources for reproducible geoscience benchmarking;

\item \textbf{Facies}: the Kaggle Well Log Facies Dataset\footnote{\url{https://www.kaggle.com/datasets/imeintanis/well-log-facies-dataset}}, where our benchmark split uses seven wells from the Council Grove gas reservoir (Kansas) with facies labels derived from core observations;

\item \textbf{FORCE}: the Kaggle Well logs dataset for machine learning\footnote{\url{https://www.kaggle.com/datasets/faresazzam/well-logs-dataset-for-machine-learning}}, released for lithology (rock-type) prediction from multiple geophysical well-log measurements and associated with the FORCE 2020 lithology prediction context;

\item \textbf{GeoLink}: the GEOLINK-S2 well-log dataset\footnote{\url{https://github.com/LukasMosser/geolink_dataset?tab=readme-ov-file}} accessed via the geolink\_dataset repository, which provides analysis notebooks and preprocessing code for the GEOLINK-S2 data.
\end{itemize}

\begin{table}[b]
    \centering
    \caption{Detailed specification of input features and target classes for the four benchmark datasets. The input features represent the exact channels used by GeoDecider.}
    \label{tab:dataset_features}
    \small
    \renewcommand{\arraystretch}{1}
    \renewcommand{\tabularxcolumn}[1]{m{#1}}
    \begin{tabularx}{\linewidth}{l c >{\raggedright\arraybackslash}X l}
        \toprule
        \textbf{Dataset} & \textbf{\# Cls} & \textbf{Input Features (Curves)} & \textbf{Target} \\
        \midrule
        \textbf{SEAM} & 7 & \textbf{9 Channels:} Depth, Bed Dip ($X, Y$), Total Porosity, Horizontal Resistivity, TTI Dip ($X, Y$), P-wave Velocity ($V_p$), S-wave Velocity ($V_s$) & Layers \\
        \midrule
        \textbf{Facies} & 9 & \textbf{7 Channels:} Depth, Gamma Ray (GR), Deep Induction Resistivity ($\log_{10}$), Neutron-Density Porosity Diff ($\Delta\phi$), Neutron-Density Porosity (PHIND), Photoelectric Effect (PE), Nonmarine-Marine Indicator & Lithofacies \\
        \midrule
        \textbf{FORCE} & 5 & \textbf{9 Channels:} Depth, Caliper (CALI), Sonic Slowness (DTC), Gamma Ray (GR), Neutron Porosity (NPHI), Bulk Density (RHOB), Coordinates ($X, Y, Z$) & Lithology \\
        \midrule
        \textbf{GeoLink} & 11 & \textbf{9 Channels:} Depth, Caliper (CALI), Neutron Porosity (NPHI), Bulk Density (RHOB), Gamma Ray (GR), Sonic Slowness (DTC), Resistivity (Deep, Shallow, Medium) & Lithology \\
        \bottomrule
    \end{tabularx}
\end{table}

\begin{table}[t]
\centering
\caption{Well-level split statistics for the benchmark datasets.}
\label{tab:dataset_split_statistics}
\small
\renewcommand{\arraystretch}{1.05}
\begin{tabular}{l|cccc}
\toprule
\textbf{Statistics} & \textbf{SEAM} & \textbf{Facies} & \textbf{FORCE} & \textbf{GeoLink} \\
\midrule
\# of total wells & 5 & 7 & 11 & 128 \\
\# of wells in training set & 3 & 4 & 4 & 80 \\
\# of wells in validation set & 1 & 1 & 2 & 23 \\
\# of wells in test set & 1 & 2 & 5 & 25 \\
\# of total samples & 7,092 & 3,164 & 52,766 & 580,205 \\
Sampling interval (m) & 10 & 0.5 & 0.15 & 0.125 \\
\bottomrule
\end{tabular}
\end{table}

\section{Dataset Splits and Leakage Control}
\label{app:split_protocol}
All benchmark datasets are partitioned at the well level, so wells used for testing are disjoint from those used for training and validation. Table~\ref{tab:dataset_split_statistics} summarizes the number of wells assigned to each split. Preprocessing statistics, including normalization and imputation values, are estimated from the training wells and then applied to validation and test wells. The routing threshold $\tau$ is selected on validation wells before test inference. The neighbor retrieval index is built only from labeled training wells, and test-well labels are never exposed to the retrieval tool, Evidence Profile construction, LLM prompts, evidence reconciliation, or geological refinement. 

\section{Baselines}
\label{Appendix_baseline}
We compare GeoDecider with representative baselines from four categories: traditional machine learning methods, deep time-series classification models, LLM-based approaches, and time-series foundation models. These baselines cover different paradigms ranging from feature-based prediction and neural temporal modeling to large-scale pretrained models, providing a comprehensive evaluation of the proposed evidence-grounded reasoning framework.
\begin{itemize}
\item \textbf{XGBoost}:
XGBoost is a tree-based gradient boosting method that builds an ensemble of decision trees to optimize a differentiable loss function. It is widely used as a strong traditional machine learning baseline due to its robustness and efficiency on tabular and time-series features.

\item \textbf{nn-DTW}:
nn-DTW combines nearest-neighbor classification with Dynamic Time Warping distance to handle temporal misalignments between time series. It serves as a classical non-parametric baseline that focuses on shape similarity rather than learned representations.

\item \textbf{GBDT}:
Gradient Boosting Decision Trees (GBDT) iteratively fits shallow decision trees to the residuals of previous trees to minimize prediction error. Compared with XGBoost, it provides a simpler yet competitive boosting baseline for sequence-level classification on hand-crafted features.

\item \textbf{LSTMFCN}:
LSTMFCN integrates a Long Short-Term Memory (LSTM) branch with a Fully Convolutional Network (FCN) branch to jointly capture temporal dependencies and local patterns in time series. This architecture has become a common deep learning baseline for multivariate time-series classification tasks.

\item \textbf{MLP}:
The Multilayer Perceptron (MLP) treats each input segment as a fixed-length feature vector and performs classification through stacked fully-connected layers with nonlinear activations. Despite its simplicity and lack of explicit temporal modeling, it offers a strong baseline when features are informative.

\item \textbf{MiniRocket}:
MiniRocket applies a fixed set of convolutional kernels to time series and uses simple summary statistics as features, followed by a linear classifier. It is designed to provide very fast and competitive time-series classification with minimal training overhead.

\item \textbf{InceptionTime}:
InceptionTime is a deep convolutional architecture for time-series classification that uses Inception-style multi-scale convolutional blocks. By capturing patterns at different temporal resolutions, it achieves strong accuracy on a wide range of benchmark datasets.

\item \textbf{InstructTime}:
InstructTime reformulates time-series classification as an instruction-following task for multimodal language models. It leverages natural-language descriptions of tasks and features to enable flexible, training-free or lightly fine-tuned time-series understanding.

\item \textbf{TableTime}:
TableTime converts time series into tabular formats and uses large language models' table-understanding capabilities for classification. This approach treats time series as structured tables, enabling training-free inference with generic LLMs.

\item \textbf{GPT4TS}:
GPT4TS is a time-series analysis framework that adapts pretrained language-model-style architectures to forecasting and classification. It views time series as token sequences and leverages generative pretraining for downstream time-series tasks.

\item \textbf{UniTS}:
UniTS is a unified multi-task time-series model designed to support  tasks such as classification, forecasting, and imputation within a  backbone. It serves as a strong foundation model baseline for evaluating specialized methods.

\item \textbf{MOMENT}:
MOMENT is an open family of time-series foundation models pretrained on large-scale heterogeneous time-series corpora. It provides general-purpose representations that can be adapted to downstream classification tasks with minimal task-specific tuning.
\end{itemize}

\begin{table}[b]
\centering
\caption{Performance comparison with different base LLMs.}
\renewcommand{\arraystretch}{1.1}
\label{comparison}
\resizebox{\columnwidth}{!}{
\begin{tabular}{lcccccc}
\toprule
\multirow{2}{*}{Base LLM} & \multicolumn{3}{c}{SEAM} & \multicolumn{3}{c}{Facies} \\ \cmidrule(lr){2-4} \cmidrule(lr){5-7}
 & Precision & Recall & F1 & Precision & Recall & F1 \\ \midrule
DeepSeek-R1 & \textbf{0.8628} & \textbf{0.7748} & \underline{0.7803} & \underline{0.4758} & \textbf{0.4461} & \textbf{0.4481} \\
GPT-5 & 0.8037 & \underline{0.7633} & 0.7716 & 0.4645 & 0.4284 & 0.4334 \\
Gemini 3 Pro & \underline{0.8486} & 0.7626 & \textbf{0.7841} & \textbf{0.4938} & \underline{0.4388} & \underline{0.4476} \\ \bottomrule
\end{tabular}
}
\end{table}

\section{Additional Hyperparameter Analyses}
\label{app:hyperparameter}

\subsection{Impact of Different LLM Backbones}
Table~\ref{comparison} isolates the effect of replacing only the reasoning backbone. We evaluate three LLM backbones: DeepSeek-R1, GPT-5, and Gemini 3 Pro within our framework, while the main results in Table~\ref{main_results} use DeepSeek-R1 as the default backbone. As shown in Table~\ref{comparison}, the backbones achieve broadly comparable precision, recall, and weighted F1 on SEAM and Facies, with modest differences. This suggests that performance gains mainly arise from the system-level design rather than the specific LLM backbone, demonstrating that the proposed framework is largely backbone-agnostic and flexible for deployment under different constraints.

\subsection{Effect of Temperature on Model Performance}
\begin{figure}[b]
    \centering
    \includegraphics[width=\linewidth]{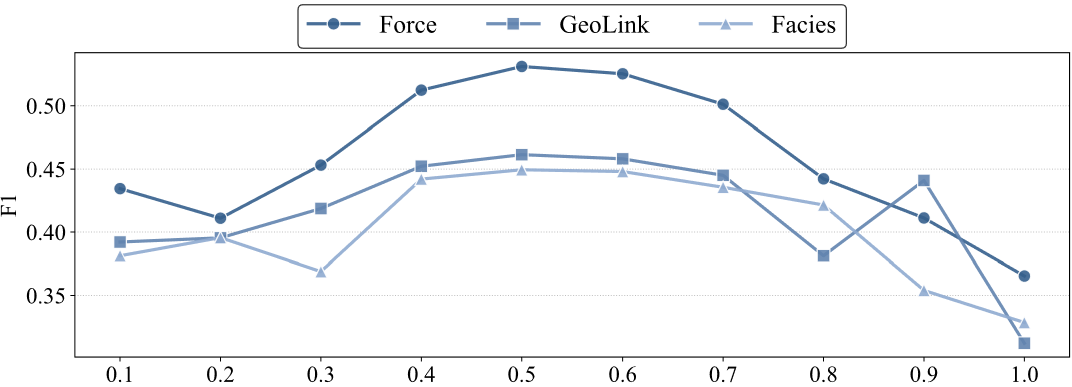}
    \caption{Sensitivity analysis of GeoDecider to the temperature parameter of LLMs, evaluated by weighted F1 score.}
    \label{wendu}
    \Description{Line chart showing GeoDecider weighted F1 sensitivity under different LLM temperature settings.}
\end{figure}

Figure~\ref{wendu} shows the sensitivity of GeoDecider to the temperature parameter. As the temperature increases, the weighted F1 score improves on the evaluated datasets, peaking at a moderate temperature. This indicates that an appropriate temperature balances exploration and determinism, enhancing multi-perspective evidence integration. However, further increases lead to performance degradation, suggesting that overly stochastic reasoning reduces stability and harms prediction quality. Overall, GeoDecider exhibits sensitivity to temperature, with an optimal range for robust performance.

\end{document}